%% file: main.tex
\documentclass[letterpaper, 10 pt, journal, twoside]{IEEEtran}
\IEEEoverridecommandlockouts

\usepackage{times}
\usepackage{makecell}

\usepackage{etoolbox}
\makeatletter
\patchcmd{\@makecaption}
  {\scshape}
  {}
  {}
  {}
\makeatother

\usepackage{tikz}

\usepackage[numbers]{natbib}
\usepackage{multicol}
\usepackage{multirow}
\usepackage{graphicx}
\usepackage[bookmarks=true]{hyperref}
\usepackage{amsmath}
\usepackage{amssymb}
\usepackage{float}
\usepackage{pifont}
\usepackage{placeins}
\usepackage{mathdots}
\usepackage[caption=false,font=normalsize,labelfont=sf,textfont=sf]{subfig}
\usepackage{xspace}
\usepackage{svg}
\usepackage{rotating}
\usepackage{fontawesome}
\usepackage{pifont}
\usepackage{bm}

\newcommand{\algname}{STAR-VLM\xspace}

\usepackage[table]{xcolor}

\begin{document}

% \title{LiHi-GS: \textbf{Li}DAR-Supervised Gaussian Splatting for \textbf{Hi}ghway Driving Scene Reconstruction}

\title{\algname: Spatiotemporal Grounding Vision-Language Models for Motion and Velocity Estimation via Automotive Radar Supervision}

\author{
Pou-Chun Kung,
Aryaman Rao,
Utkrisht Sahai,
Hemanth Murali,
Yi Liu,
Rui-Yu Lin,
Katherine A. Skinner

% \thanks{Manuscript received: April 6, 2025; Revised June 23, 2025; Accepted October 2, 2025.} %Use only for final RAL version
% \thanks{This paper was recommended for publication by Editor Pascal Vasseur upon evaluation of the Associate Editor and Reviewers' comments.}
% \thanks{$^1$X. Zhang and N. Jaipuria are with Latitude AI. \texttt{\{xzhang, njaipuria\}@lat.ai}.}
\thanks{All authors are with the University of Michigan, Ann Arbor, MI 48109. \texttt{\{pckung, aryamanr, usahai, hems, yiliuhh, imrui, kskin\}@umich.edu}.}
% \thanks{$^\dagger$ The project was carried out during P. Kung's internship at Latitude AI.}
% \thanks{Digital Object Identifier (DOI): see top of this page.}
}

% \markboth{IEEE Robotics and Automation Letters. Preprint Version. Accepted October, 2025}{Kung \MakeLowercase{\textit{et al.}}: LiHi-GS: \textbf{Li}DAR-Supervised Gaussian Splatting for \textbf{Hi}ghway Driving Scene Reconstruction} 

\maketitle

\begin{abstract}
Vision-language models (VLMs) are emerging as a key component of embodied intelligence, with growing applications in auto-labeling and end-to-end autonomous driving. However, existing approaches for improving spatiotemporal reasoning in VLMs often rely on complex preprocessing pipelines, expensive human annotations, or synthetic data, which limit scalability and introduce potential sim-to-real gaps. Moreover, although these methods have improved spatiotemporal understanding, they still lack strong metric reasoning capabilities for dynamic scenes, such as estimating object motion in real-world units. Prior work has explored LiDAR-based metric depth supervision to enhance spatial perception, but it does not directly address temporal reasoning.
We introduce \algname, an automotive radar-supervised framework that enhances spatiotemporal VLMs with motion reasoning and metric velocity estimation for autonomous driving. Automotive radar is a low-cost and widely deployed sensor that provides complementary spatiotemporal supervision through range and Doppler measurements. By leveraging these measurements as label-free ground truth during training, \algname improves the metric spatiotemporal reasoning ability of VLMs. 
Through experiments on driving scenarios, we show that \algname achieves state-of-the-art performance on both motion classification and metric velocity estimation, outperforming even task-specific methods designed for each task. These results highlight automotive radar as a scalable and cost-effective source of supervision for building metric-aware spatiotemporal VLMs for real-world autonomous driving.

\end{abstract}


\begin{figure}
    \centering
    \includegraphics[width=1\linewidth]{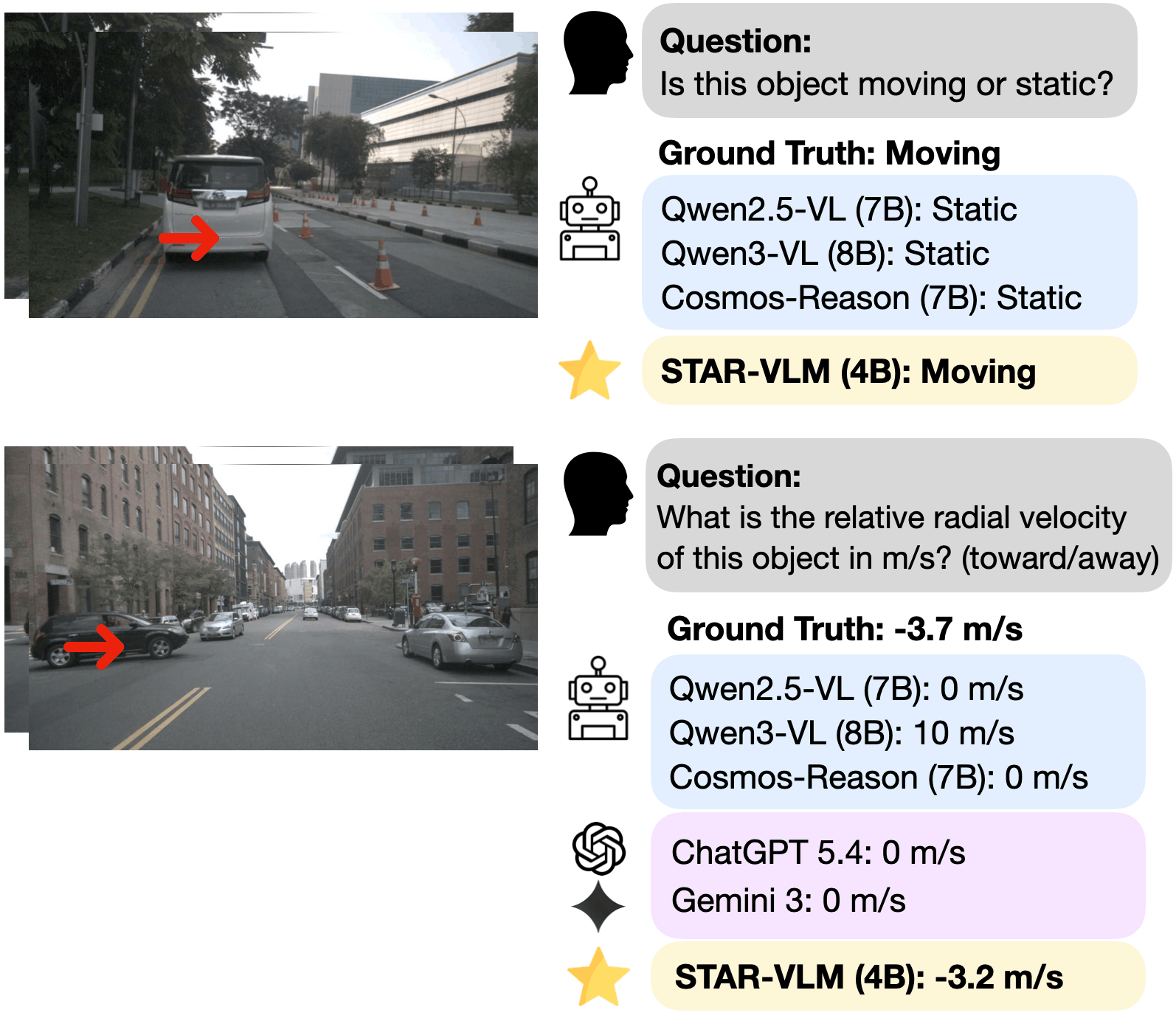}
    \caption{
    We propose \algname, a simple yet effective method that leverages automotive radar data to improve the spatiotemporal reasoning capabilities of VLMs for autonomous driving. Existing open-source VLMs struggle with temporal motion classification, and even the latest proprietary VLMs, such as GPT-5.4 and Gemini-3, still face challenges in estimating the metric velocity of moving objects.
    }
    \label{fig:teaser}
\end{figure}

\section{Introduction}

Vision-language models (VLMs) are rapidly emerging as a foundation for embodied intelligence, with growing relevance to applications such as vision-language-action (VLA) and end-to-end autonomous driving~\cite{simlingo, autovla}. By coupling visual perception with language-based reasoning, VLMs offer a promising path toward general-purpose systems that can interpret complex scenes, follow high-level instructions, and support interpretable human-like downstream decision-making. In driving scenarios, however, this requires more than understanding static scenes. Autonomous systems operate in dynamic environments populated by moving agents and safety-critical interactions, making spatiotemporal understanding a fundamental requirement. 

Recent work has begun extending VLMs from static visual understanding to spatiotemporal reasoning over videos and driving scenes. These methods aim to improve a model’s ability to reason about object motion, temporal events, and scene dynamics, which are essential for embodied decision-making. However, many spatiotemporal VLMs rely on complex preprocessing pipelines, costly human annotations, or synthetic supervision, which limit scalability and raise concerns about sim-to-real transfer~\cite{spatialvlm, robospatial, vlm4d}.

Beyond scalability, prior methods also lack strong metric reasoning, especially for dynamic scene understanding. In driving scenarios, estimating distance and motion in physically meaningful units is important for auto-labeling and downstream driving tasks. Recent work~\cite{depthlm} has explored LiDAR or depth-based supervision to improve the metric spatial perception of VLMs without complex data preprocessing and human labeling. The method, DepthLM, shows that external sensor supervision can help VLMs learn stronger metric grounding. However, such supervision primarily addresses spatial perception and does not directly provide temporal motion cues. Figure~\ref{fig:teaser} shows that metric spatiotemporal reasoning remains underexplored in current VLMs.

\begin{figure}
    \centering
    \includegraphics[width=1\linewidth]{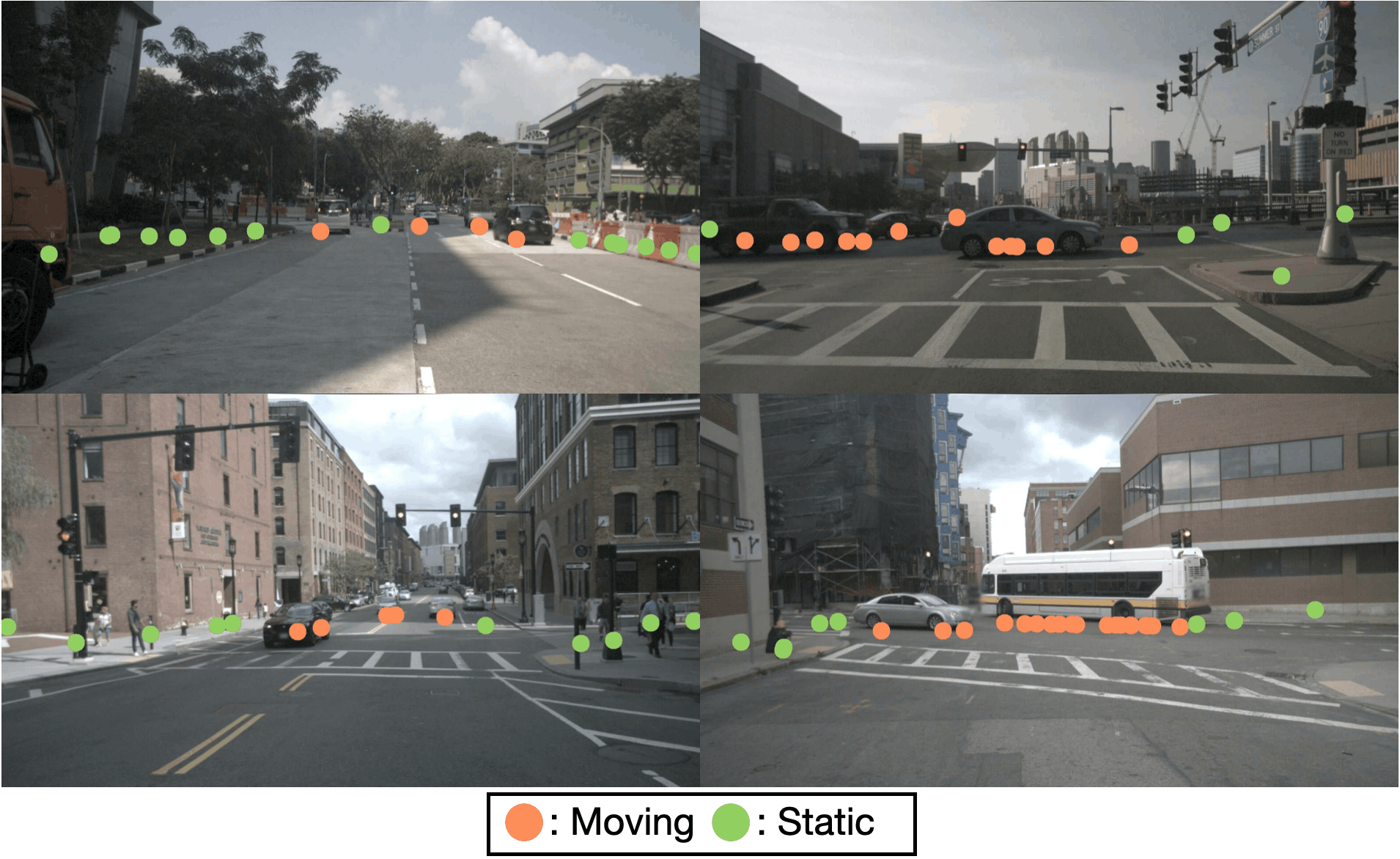}
    \vspace{-0.17in}
    \caption{Beyond sparse point clouds, the automotive radar provides point-wise motion-state estimates derived from Doppler measurements. }
    \label{fig:mc}
\end{figure}

\begin{figure}
    \centering
    \includegraphics[width=1\linewidth]{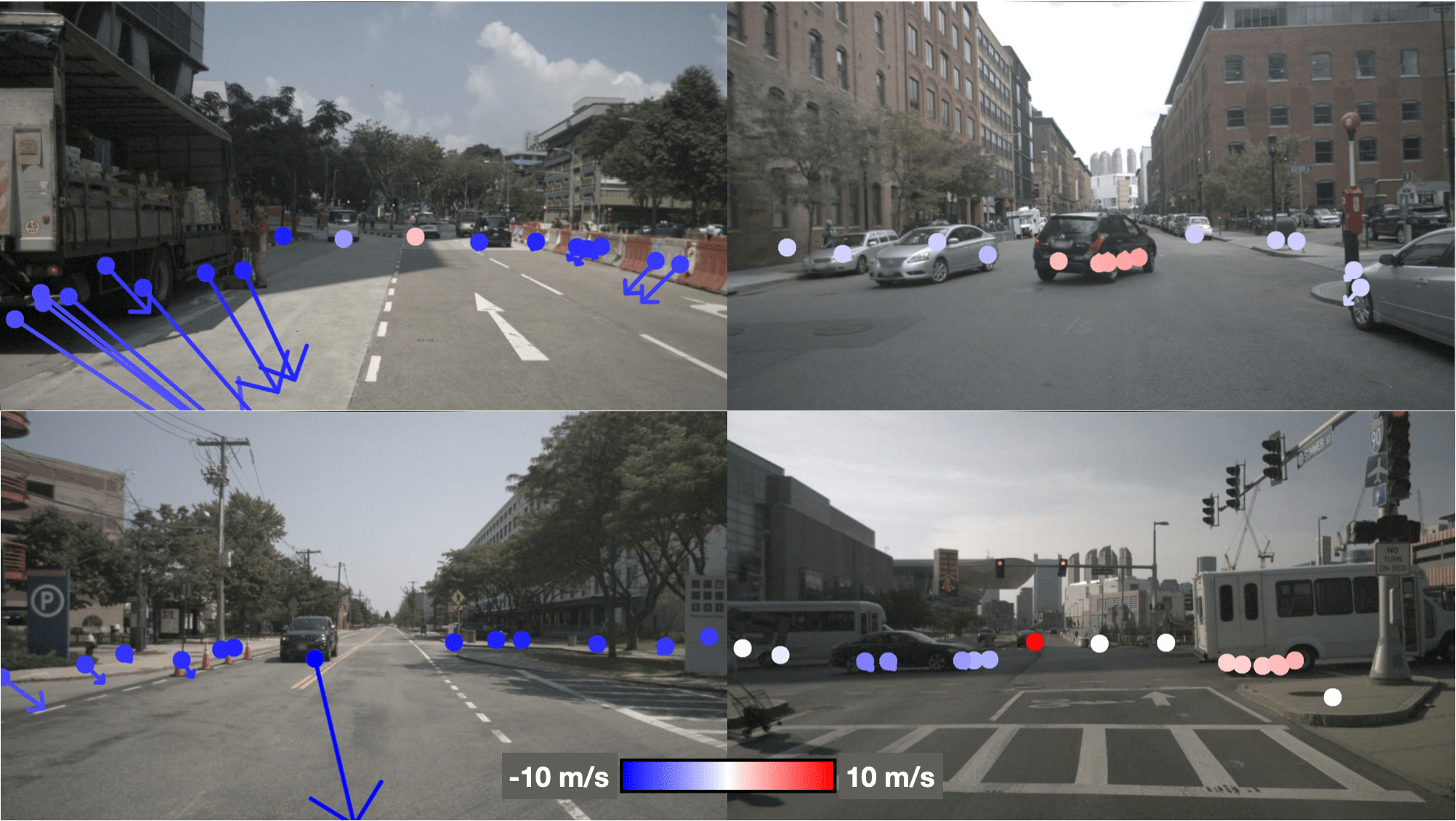}
    \caption{Automotive radar provides relative radial velocity, indicating whether an object is moving toward or away from the sensor, based on measured Doppler information.}
    \label{fig:radvel}
\end{figure}

In this work, we argue that radar is a natural supervision source for metric spatiotemporal learning in autonomous driving. Radar is low cost, widely deployed on automotive platforms, naturally aligned with the driving-relevant plane, and capable of directly measuring scene motion. In particular, radar provides Doppler measurements for temporal supervision, making it especially well-suited for improving motion reasoning. Figures~\ref{fig:mc} and~\ref{fig:radvel} illustrate motion state and radial velocity measurements from radar, respectively. Compared with manual annotations or simulator-generated supervision, radar offers a scalable, real-world source of physically grounded training signals.

Motivated by this observation, we introduce \algname, a radar-supervised framework for enhancing spatiotemporal VLMs with metric scene understanding. \algname leverages radar signals during training to improve both spatial grounding and temporal motion reasoning in VLMs. By using radar range and Doppler cues as supervision, our framework equips VLMs with a stronger metric understanding of dynamic driving scenes, particularly for driving-relevant objects such as surrounding vehicles.

We evaluate \algname on motion understanding tasks in driving environments and show that it achieves state-of-the-art performance on both motion classification and metric velocity estimation. Notably, \algname outperforms even task-specific methods tailored to these tasks, highlighting the effectiveness of radar supervision for physically grounded scene understanding. Our results suggest that radar is a powerful and scalable supervision source for building metric-aware spatiotemporal VLMs, as shown in Figure~\ref{fig:teaser}. Altogether, the contributions of this paper are as follows:
\begin{itemize}
    \item We propose \algname, a radar-supervised framework that leverages Doppler signals to enhance spatiotemporal reasoning in VLMs, with a particular focus on the underexplored problem of metric velocity reasoning in VLMs.
    % \item We achieve state-of-the-art VLM on both motion classification and metric radial velocity estimation on our self-designed and a public temporal VLM benchmark.
    % \item We show that \algname even outperforms existing task-specific pure vision methods on both motion classification and object radial velocity estimation in driving scenes.
    \item We demonstrate that \algname achieves state-of-the-art VLM performance on both motion classification and metric radial velocity estimation across our self-designed benchmark and a public temporal VLM benchmark.
    \item We further show that \algname outperforms existing task-specific language-free models on both motion classification and object-level radial velocity estimation in driving scenes.
\end{itemize}

\section{Related Work}

\subsection{VLMs for Embodied Intelligence}
VLMs have emerged as an important foundation for embodied intelligence, where language-conditioned perception and reasoning support downstream control and decision-making. In robotics, RT-2 shows that vision-language pretraining can be transferred to embodied control by representing actions as text tokens, establishing a strong connection between VLMs and VLA systems~\cite{rt2}. In autonomous driving, recent works such as DriveLM, DriveVLM, and DriveMLM extend VLMs to scene understanding, reasoning, and planning, highlighting their potential for interactive and interpretable driving systems~\cite{drivelm, drivevlm, drivemlm}. More recently, autonomous driving VLA systems have adopted VLMs as their backbone and directly output driving trajectories~\cite{simlingo, autovla}.
% todo: add autoVLA, alpamayo

\subsection{Spatiotemporal VLMs for Driving}
Recent efforts have pushed driving VLMs beyond single-image reasoning toward spatiotemporal understanding. DriveMLLM introduces a benchmark for evaluating spatial understanding in autonomous driving \cite{drivemllm}. FutureSightDrive further explores spatiotemporal chain-of-thought reasoning for end-to-end driving \cite{futuresightdrive}, STSBench benchmarks holistic spatiotemporal reasoning over traffic scenarios using multi-view and temporal context \cite{stsbench}, and TADBench focuses specifically on temporal understanding in autonomous driving with nearly 6,000 question-answer (QA) pairs \cite{tadbench}. At the same time, DriveBench shows that current driving VLMs can produce plausible answers without robust visual grounding \cite{drivebench}. 
%Both DriveBench and TADBench provide benchmark settings for evaluating temporal understanding in autonomous-driving VLMs \cite{drivebench, tadbench}. 
% We show that \algname, with radar supervision, outperforms existing models on temporal understanding tasks on TADBench. %both DriveBench and TADBench.
% todo: SpatialVLM

\begin{figure*}
    \centering
    \includegraphics[width=1.0\linewidth]{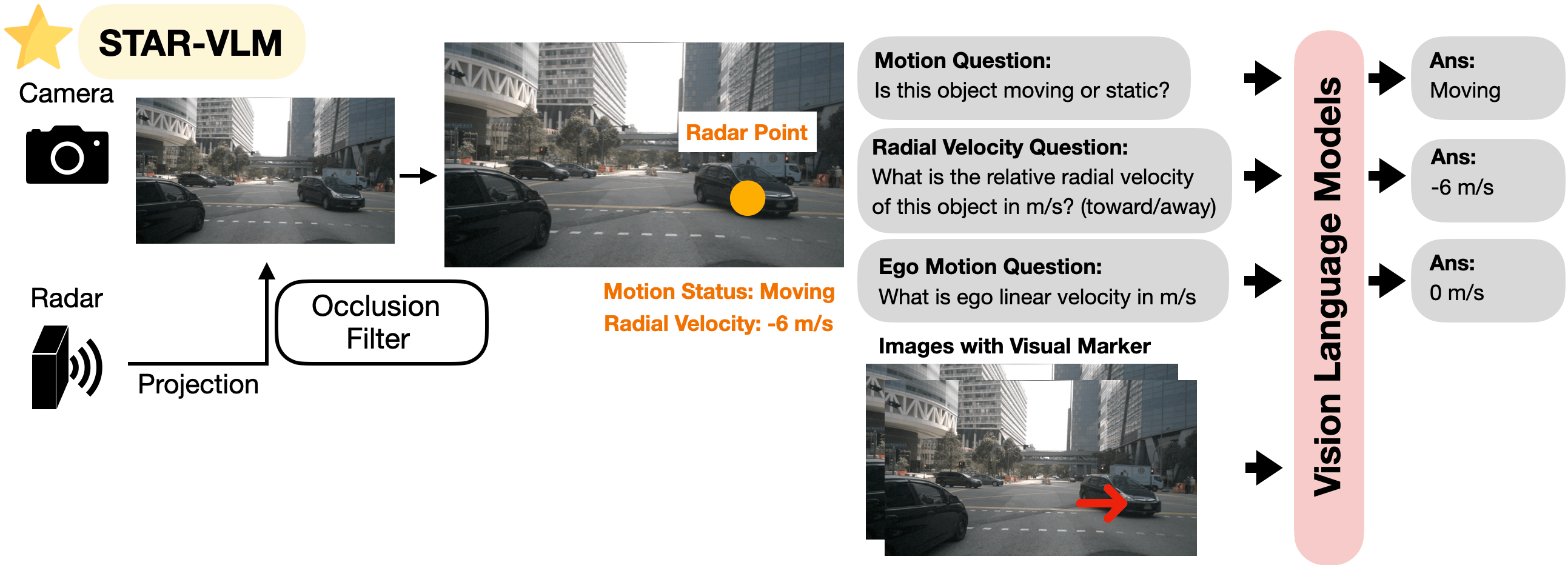}
    \caption{Overview of \algname. We project radar points onto the camera image to obtain supervision labels and construct a QA dataset from radar measurements and ego-motion signals for VLM supervision. Each queried radar point is highlighted with a visual marker as a visual prompt.}
    \label{fig:overview}
\end{figure*}

\subsection{Metric Reasoning in VLMs}
A complementary line of research focuses on endowing VLMs with stronger metric and geometric understanding. In particular, DepthLM demonstrates that standard VLMs can be adapted to perform metric depth estimation, indicating that explicit supervision can significantly enhance geometric grounding without the need for task-specific architectures~\cite{depthlm}. However, existing efforts mainly address the spatial dimension of metric reasoning. Metric temporal reasoning, such as estimating object velocity in metric space, remains largely unexplored in VLMs, particularly in autonomous driving scenarios.

% \subsection{Radar for autonomous driving and motion understanding.}
% Radar is attractive for autonomous driving because it is widely deployed in automotive platforms and directly provides motion-sensitive measurements through Doppler in addition to range information. Recent surveys on 4D millimeter-wave radar and radar for autonomous driving highlight its growing role in perception and autonomous driving, including object detection, tracking, localization, and motion estimation \cite{4dmmwaveradar, radarsforautonomousdriving, exploringradardatarepresentations}. Prior radar-specific methods have leveraged these properties for velocity estimation and scene flow: for example, \cite{selfsupervisedvelocityestimation} learns Cartesian object velocity from automotive radar, while \cite{hiddengems} uses cross-modal supervision for 4D radar scene flow estimation. More recently, TARS improves radar scene flow estimation by incorporating traffic-level motion structure \cite{tars}. These works demonstrate the strength of radar for motion understanding, but they are designed as specialized perception models rather than as supervision sources for improving the reasoning capabilities of VLMs.

\subsection{Radar Cross-Modal Fusion and Supervision}
Recent work has explored radar in autonomous driving through radar-camera fusion for detection, depth estimation, and velocity estimation. For detection, methods such as CenterFusion, RCBEVDet, and CRKD show that radar provides complementary geometric and motion cues for 3D detection and tracking \cite{centerfusion, rcbevdet, crkd}. Radar has also been combined with monocular images for depth estimation, where prior work uses radar to improve metric spatial geometry through direct fusion or depth completion \cite{radardepth, tacodepth, mmwaveradarcameradepth}. In addition, radar-camera methods have been developed for velocity estimation by combining tangential velocity from visual motion cues with radial velocity from radar Doppler measurements \cite{fullvelocityradarreturns, pow4r}.

More closely related to our setting, several works use radar as supervision during training instead of as a required sensing modality at inference. For example, DoGFlow and RaLiFlow use radar-derived motion labels to improve scene flow estimation, while R4Dyn and Radar as a Teacher use radar as weak supervision for monocular depth estimation and image-based detection, respectively \cite{dogflow, raliflow, r4dyn, radarasateacher}. In contrast, our work uses radar supervision to improve metric temporal reasoning in a spatiotemporal VLMs.

% [radar+camera det.]
% CenterFusion
% RCBEVDet
% CRKD
% CR3DT (+tracking)

% [radar+camera depth est.]
% Depth Estimation from Monocular Images and Sparse Radar Data -> camera+radar depth estimation
% TacoDepth
% Depth Estimation Based on MMwave Radar and Camera Fusion
% Sparse Beats Dense

% [radar+camera velocity est.]
% Full-Velocity Radar Returns by Radar-Camera Fusion
% POW4R

% [radar-supervision]
% DoGFlow: radar as motion labeler to improve lidar scene flow
% RaLiFlow: radar as motion labeler to improve lidar scene flow

% R4Dyn: radar weakly-supervised monodepth estimation
% Radar as a Teacher: radar weakly-supervised camera detection

\begin{figure*}
    \centering
    \includegraphics[width=0.9\linewidth]{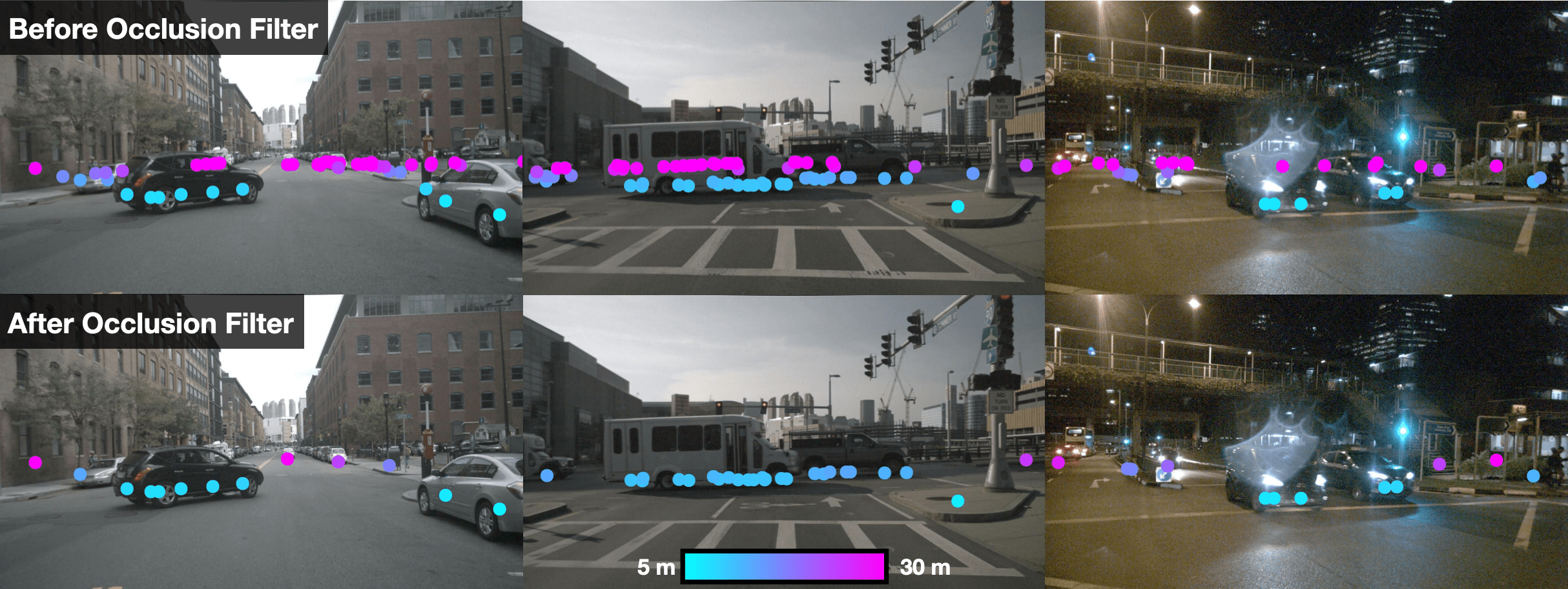}
    \caption{Projection of radar points into image space with and without the proposed occlusion filter. Point color represents the distance of each radar point. Without occlusion filtering, radar points from objects hidden behind visible surfaces are still projected into the image, which can introduce incorrect training labels. After applying the proposed occlusion filter, points from occluded objects are filtered out. }
    \label{fig:occlusion_filter}
\end{figure*}

\subsection{Camera Motion Classification and Velocity Estimation}
A substantial body of work studies motion understanding from monocular video, spanning motion segmentation, object-level velocity estimation, and dense pixel-level motion estimation. For motion segmentation, prior methods such as MoSeg, M3Former, and SegAnyMo aim to separate moving objects from background or ego-motion in monocular videos by combining appearance, motion, geometric, or tracking cues \cite{moseg, m3former, seganymotion}. These methods provide strong baselines for identifying dynamic regions, but they primarily focus on segmentation rather than quantitative motion estimation.

For velocity estimation at the object level, camera-only methods have also been developed to estimate vehicle motion and velocity. \cite{camerabasedvehiclevelocityestimation} studies relative vehicle velocity prediction directly from monocular video, while \cite{kinematic3dobjectdetection} leverages temporal kinematic cues to improve monocular 3D detection and additionally predicts object velocity. These methods demonstrate that monocular video can support object-level motion estimation, but they are designed as task-specific pipelines rather than general-purpose spatiotemporal reasoning VLMs.

For velocity estimation at the pixel level, monocular scene flow methods estimate dense 3D motion from image sequences. \cite{selfsupervisedmonocularsceneflow} is a representative self-supervised approach for jointly recovering geometry and motion from monocular video, while more recent methods such as Any4D\cite{any4d} and ZeroMSF~\cite{zeromsf} further improve dense metric 4D reconstruction and zero-shot monocular scene flow estimation. These methods are closely related to our evaluation setting because they recover physically meaningful motion from camera observations. However, despite their dense predictions, they often struggle to provide accurate metric velocity estimates. In contrast, our work improves metric velocity estimation by leveraging a powerful VLM pretrained on large-scale data.

% Motion segmentation / moving object segmentation from monocular video:
% MoSeg
% M3Former
% SegAnyMo
% RoMo

% Object-level velocity estimation from monocular video:
% Camera-based vehicle velocity estimation from monocular video
% Kinematic 3D Object Detection in Monocular Video

% pixel velocity estimation from monocular video:
% Self-Supervised Monocular Scene Flow Estimation
% Any4D
% ZeroMSF

\section{Method}
The overview of \algname is shown in Figure~\ref{fig:overview}. To enhance the spatiotemporal reasoning capability of VLMs using automotive radar, and to systematically evaluate their spatiotemporal understanding, we preprocess radar point clouds and human-labeled annotations to construct QA pairs for both training and evaluation. We describe the design of the training data and benchmarks in Sec.~\ref{sec:method_bench}. To investigate effective prompt design for temporal understanding, we present the prompt formulations and input formats in Sec.~\ref{sec:method_prompt}. The training strategy is detailed in Sec.~\ref{sec:method_training}. Finally, the complete training pipeline for the full model is described in Sec.~\ref{sec:final_method}.

\subsection{STAR-Bench}
\label{sec:method_bench}
% nuscenes

% Radar QA:
% Occlusion Filter
% Anno QA:
% BBox Occlusion Filter
% motion label + radial vel

We curate \textbf{STAR-Bench} by leveraging the public nuScenes dataset~\cite{nuscenes}, which contains human annotations and synchronized camera-radar measurements for autonomous driving scenes. For training, we use the front-view images and front radar data from the trainval split, excluding the mini split, resulting in approximately 34K training images. For evaluation, we use the mini split, which contains around 400 images drawn from diverse driving segments.

We construct two variants of \textbf{STAR-Bench}. \textbf{STAR-Bench-radar} is built by projecting radar measurements onto the image plane to provide point-level radial velocity, motion status, and distance cues. \textbf{STAR-Bench-anno} is built by projecting ground-truth 3D bounding boxes onto the image plane to provide object-level annotations, including full velocity, motion status, and distance.

When constructing \textbf{STAR-Bench-radar}, we observe that radar’s penetrative and long-range sensing properties often produce returns that are occluded in the camera view. Directly using these points can introduce inaccurate supervision and evaluation labels for VLMs, as shown in Fig.~\ref{fig:occlusion_filter} (top). We therefore apply an occlusion filter to remove radar points that are likely occluded in the image, as illustrated in Fig.~\ref{fig:occlusion_filter} (bottom). Concretely, for each radar point projected onto the image plane, we search for its nearest neighbor in image coordinates. If the neighbor point is within $N = 30$ pixel distance and closer in depth by more than $\tau = 0.3$ m, the point is considered occluded and removed. 

% radar velocity & dyn_prob class

A similar issue arises for \textbf{STAR-Bench-anno}, since human annotations are provided in LiDAR space. We thus sort 3D bounding boxes from near to far and remove annotations that are occluded in the image plane with $\mathrm{IoU} > 0.3$. We use the center of the bounding box at 0.3 m height as the target point to query VLMs for both training and evaluation.

% annotation extract center point at sensor height 
% todo: add figure

\subsection{Prompt Design}
\label{sec:method_prompt}
% arrow | pixel_coord | arrow+pixel_coord
% temporal input: video vs multi-image input
% ego-motion compensation

\subsubsection{Pixel reference}
DepthLM~\cite{depthlm} observes that existing VLMs often refer to a queried pixel using its textual coordinates $(x, y)$. However, their experiments show that many VLMs struggle to reliably ground text-based coordinates to the corresponding image location, even after training. To address this issue, DepthLM proposes using a visual prompt, as illustrated in Fig.~\ref{fig:teaser}. 

\begin{figure}
    \centering
    \includegraphics[width=0.8\linewidth]{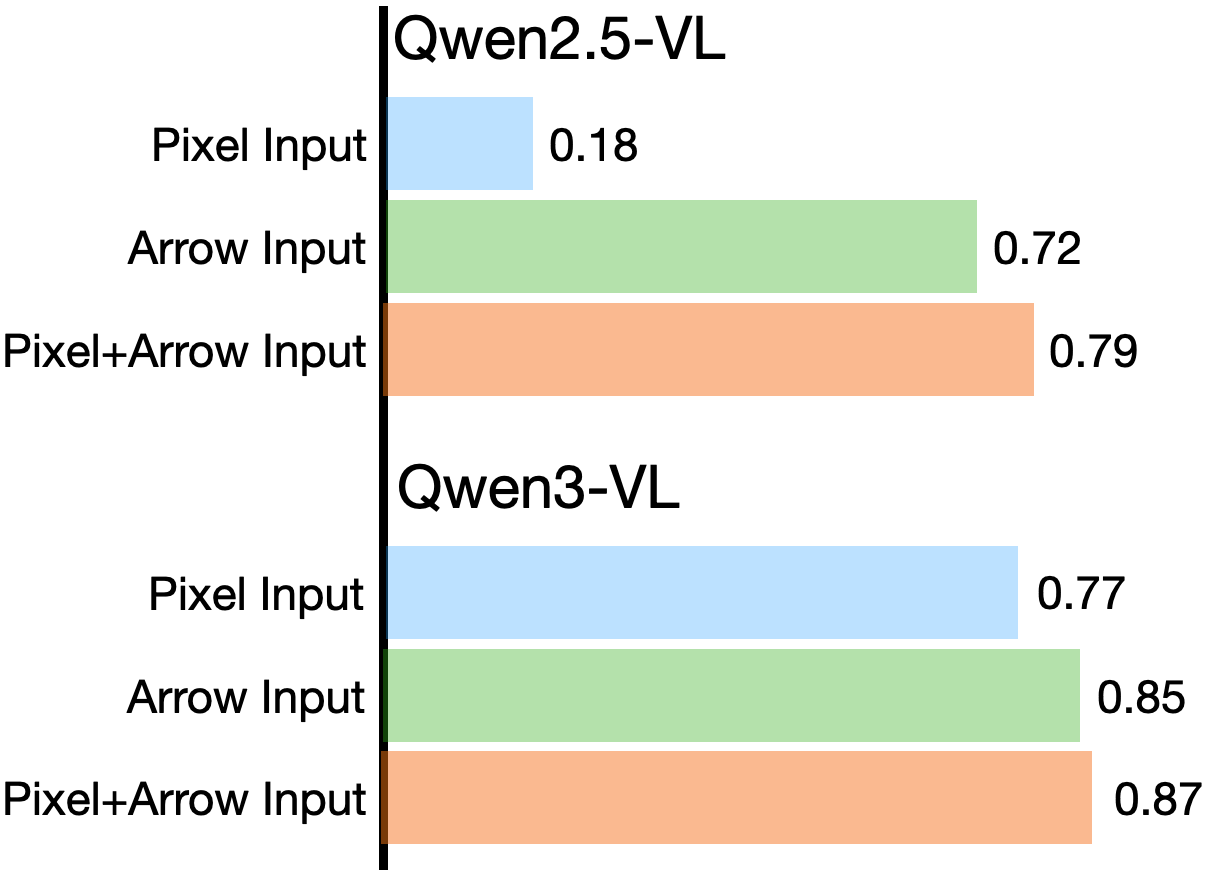}
    \caption{Impact of different input pixel references. The numbers indicate depth estimation accuracy in the DepthLM setting. Using pixel coordinates as input leads to the worst performance, whereas visual markers such as arrows perform better. Our proposed joint input, which combines pixel coordinates and visual markers, achieves the best performance across different Qwen base models.}
    \label{fig:input_prompt}
\end{figure}

In our experiments, we find that this behavior depends on the underlying base model. For example, Qwen2.5-VL is more sensitive to the choice of pixel reference and performs better when visual markers are used, whereas Qwen3-VL achieves similar performance with either textual coordinates or visual markers, as shown in Fig.~\ref{fig:input_prompt}. 

Furthermore, we find that combining both pixel coordinates and visual markers yields the best performance across different base models (Fig.~\ref{fig:input_prompt}). We hypothesize that this combination provides stronger pixel-level grounding, which not only improves the localization of the queried point but also enhances generalization across QA benchmarks that use different prompt formats.

\subsubsection{Question-answer design} For joint pixel-coordinate and visual prompting, we render a visual marker on the input image to explicitly indicate the queried pixel, and ask:
\textit{``What is the radial velocity (in m/s) of the point indicated by the red arrow at pixel $(x, y)$ relative to the camera?"}
and
\textit{``What is the motion state of the point indicated by the red arrow at pixel $(x, y)$ relative to the camera?"}
% See Appendix ... for examples of the rendered markers.

During training, we use the answer templates
\textit{``The point has motion state X."}
and
\textit{``The point has radial velocity Y m/s relative to the camera."}
We treat both X and Y as standard text outputs, where X is a binary motion label (\textit{moving} or \textit{static}) and Y is velocity rounded to two decimal places. This design avoids unnecessarily long floating-point responses while preserving sufficient precision for our 3D understanding tasks.

\subsection{Training}
\label{sec:method_training}
% training: SFT vs LoRA (improve general spatiotemporal understanding)
% SFT -> overfit
% LoRA -> preserve thinking

\subsubsection{Training method}
We follow SpatialVLM~\cite{spatialvlm} and adopt supervised fine-tuning (SFT) for training. DepthLM~\cite{depthlm} also reports that SFT achieves performance comparable to reinforcement learning, while being substantially more computationally efficient than reinforcement learning. 

% LoRA Training
% However, we observe that applying SFT to a fixed set of question-answer pairs can degrade the model’s ability to generalize across different temporal reasoning tasks. To mitigate this issue, we further explore LoRA-based adaptation. We find that LoRA achieves comparable performance on the target training tasks while better preserving the model’s reasoning ability and generalization to unseen temporal tasks. The comparison is shown in Fig.~\ref{fig:sft_vs_lora}.

\begin{figure}
    \centering
    \includegraphics[width=1\linewidth]{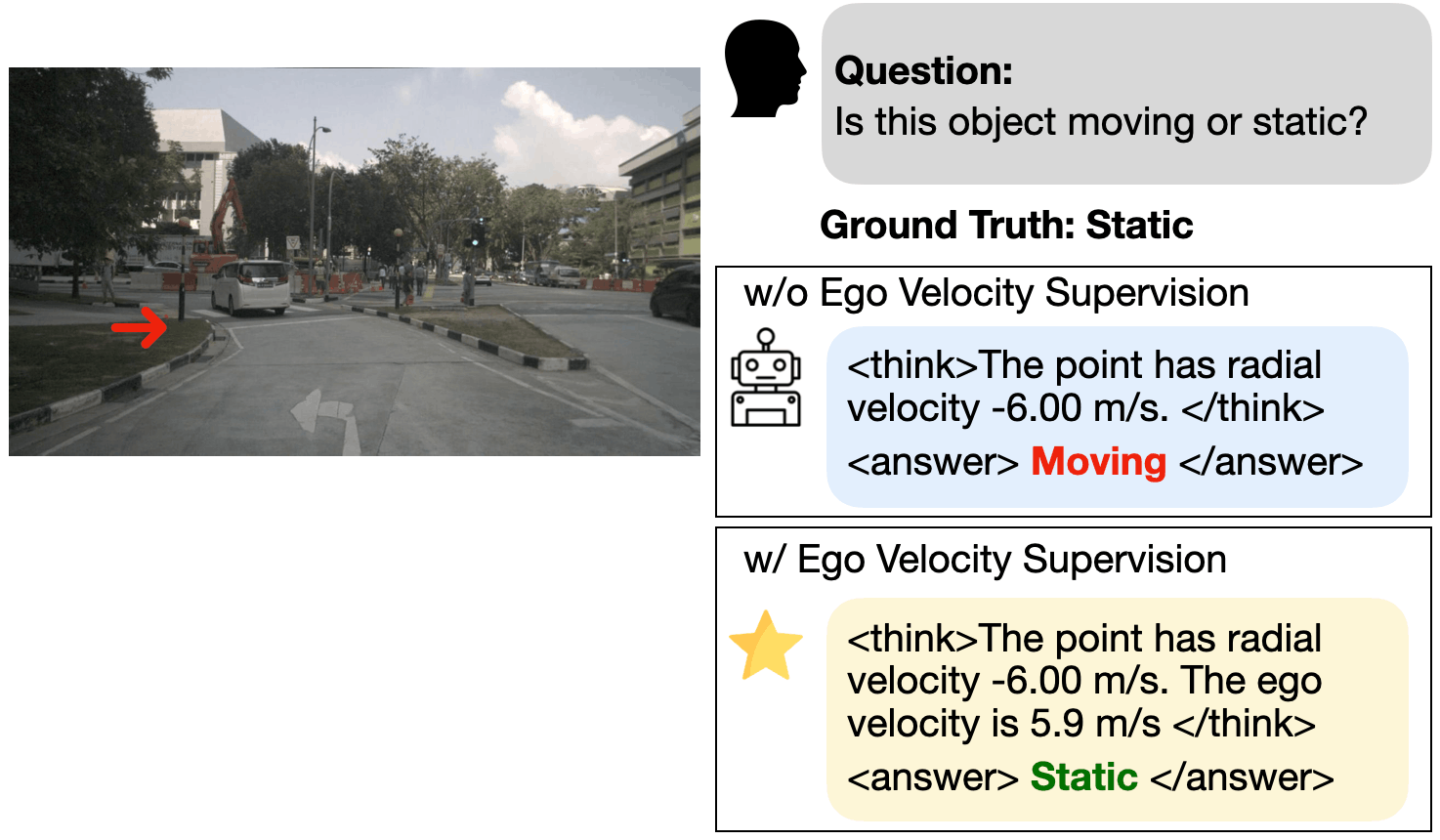}
    \caption{Illustration of the effect of ego-velocity joint training on motion classification when the model is trained only for velocity estimation. Without ego-velocity supervision, the model cannot determine whether the observed motion is caused by ego motion or object motion. With ego-velocity joint training, the model can compensate for ego motion and infer motion state in the global coordinate frame. }
    \label{fig:egovel_training}
\end{figure}

\subsubsection{Radial and ego-velocity cross-task training}
\label{sec:joint_training}
Our experiments show that radar radial velocity alone serves as an effective supervision signal for velocity inference, already enabling strong performance. Moreover, jointly training the model to predict both object radial velocity and ego velocity yields further improvements in metric velocity estimation, even with the same training data budget. We attribute these gains to the shared spatiotemporal reasoning underlying the two tasks. This joint training strategy also substantially increases the amount of metric supervision available to the model. Importantly, ego-velocity supervision can be obtained without human annotation, requiring only ego-pose estimation.

We also observe that ego-velocity joint training improves motion classification accuracy when the model is trained with radial velocity supervision. We hypothesize that ego-velocity supervision helps the model better disentangle ego motion from object motion. Figure~\ref{fig:egovel_training} presents example outputs and reasoning traces that illustrate this effect.

\subsection{Final Method}
\label{sec:final_method}

The findings from the analysis inspire us to propose the complete method of \algname. Figure~\ref{fig:overview} shows the method overview. 

\subsubsection{Pixel reference}
Unlike prior VLM-based approaches that use either textual pixel coordinates or visual markers alone~\cite{depthlm}, we provide both forms of pixel reference: pixel coordinates in the text prompt and rendered visual markers, namely small arrows pointing to the queried pixels, in the image. This joint design strengthens pixel-level grounding and improves the VLM's ability to accurately localize the queried point. After training, the model supports both visual-marker-based prompts and coordinate-based prompts, enabling greater flexibility across tasks and prompt formats.

\subsubsection{Architecture and training}
We finetune our models from pretrained VLMs without an architecture change. We apply standard SFT with the next token prediction paradigm and the cross-entropy loss on the text tokens. As suggested by \cite{depthlm}, no regression or regularization loss is needed in pure vision models. 
The \textbf{STAR-Bench-radar} data is used as the training dataset for the main model. We also trained a model with \textbf{STAR-Bench-anno} to compare the difference of training using the proposed low-cost automotive radar data and the human annotation.

\subsubsection{Resolve camera ambiguity} Following \cite{depthlm}, we also unified the camera focal length to make sure the model can scale to different cameras. 

\subsubsection{Cross-task training}
We jointly train \algname using three types of QA sets: motion classification, radial velocity estimation, and ego-velocity estimation. According to Sec.~\ref{sec:joint_training}
and the results in Sec.~\ref{sec:cross-task}, joint training with ego-velocity estimation improves both motion classification and radial velocity accuracy.

\section{Experiments}

\subsection{Implementation Details}

To simplify experiments and avoid computation overhead, we choose the 4B Qwen3-VL model~\cite{qwen3vl} as our base VLM. 

We train models with PyTorch on 34k images and $\sim$900k radar points from \textbf{STAR-Bench-radar}, i.e., 25 to 30 radar points per image, with 4 A100 GPUs for about 12 to 30 hours, depending on the model and stopping strategy. 

\subsection{Baselines}
For spatiotemporal VLM comparison, we adopt the open-source 3B and 7B versions of Qwen2.5-VL~\cite{qwen2vl} and Qwen3-VL~\cite{qwen3vl} as our primary baselines. We further compare against Cosmos-Reason~\cite{cosmos}, a physics-aware VLM for robotics and autonomous driving built upon Qwen2.5-VL with enhanced physical understanding. In addition, we include DepthLM, the first VLM capable of metric depth estimation. Specifically, DepthLM-3B~\cite{depthlm} is trained for the nuScenes dataset, while DepthLM-Pixtral is the official 12B model trained on a diverse set of datasets. We also compare against task-specific models, including state-of-the-art SegAnyMo~\cite{seganymotion} for motion classification and Any4D~\cite{any4d} for radial velocity estimation.

\subsection{Evaluation Benchmark}
To evaluate the spatiotemporal reasoning capability of VLMs for autonomous driving, we construct STAR-Bench QA sets using radar measurements and human annotations from the nuScenes dataset. We further evaluate on the public TADBench benchmark, which is designed for autonomous driving scenarios, to provide complementary evaluations of temporal understanding.

%We further evaluate on the public benchmarks TADBench and DriveBench, both of which are designed for autonomous driving scenarios, to provide complementary evaluations of temporal understanding.

\subsubsection{STAR-Bench}
The details of STAR-Bench are discussed in Sec.~\ref{sec:method_bench}. \textbf{STAR-Bench-radar} and \textbf{STAR-Bench-anno} are QA sets generated by radar measurement and human annotation separately.

\input{table/results}

\subsubsection{TADBench}
\label{sec:tadbench}
We evaluate VLMs on the motion classification QA set introduced in~\cite{tadbench}. This benchmark defines two prompt formats. The first, \textbf{Exact Visual Question-Answer (VQA)}, uses prompts such as \textit{"What best describes the motion of the vehicle in this video segment? Respond with exactly one full phrase from the following list: `Starting', `Stopping', `Turn left', `Turn right', `Change lane to the left', `Change lane to the right', `Straight, constant speed', `Stopped'."} with answers such as \textit{"Stopped"}. The second, \textbf{Multiple-Choice Question (MCQ)}, uses prompts such as \textit{"What is the motion of the vehicle in this video segment? Respond with exactly one letter corresponding to the correct option. A. Straight, constant speed B. Stopped C. Starting D. Stopping"} with answers such as \textit{"A"}.

Unlike STAR-Bench, which is used for training, TADBench does not provide explicit pixel coordinates or visual markers in the input image. Instead, the prompt refers to the target object only through natural language, such as ``the vehicle in this video segment.'' We find that \algname handles this prompt shift well and preserves strong performance under this less explicit object reference setting.

% \paragraph{DriveBench.}
% We evaluate VLMs on the question-answer set introduced in~\cite{drivebench}, focusing on the motion-status questions within the perception category. These questions are formulated as multiple-choice prompts, for example:
% \textit{"What is the moving status of the object at $(x, y)$? Please select the correct answer from the following options: A. Going Ahead B. Turn Left C. Turn Right"}
% with answers such as \textit{"A. Going Ahead"}.

% Unlike TADBench, DriveBench uses pixel-coordinate prompting to specify the target object. Since pixel coordinates are also included in our training setup, we find that \algname handles this prompt format naturally and generalizes well to the DriveBench evaluation setting.

\begin{figure}
    \centering
    \includegraphics[width=1\linewidth]{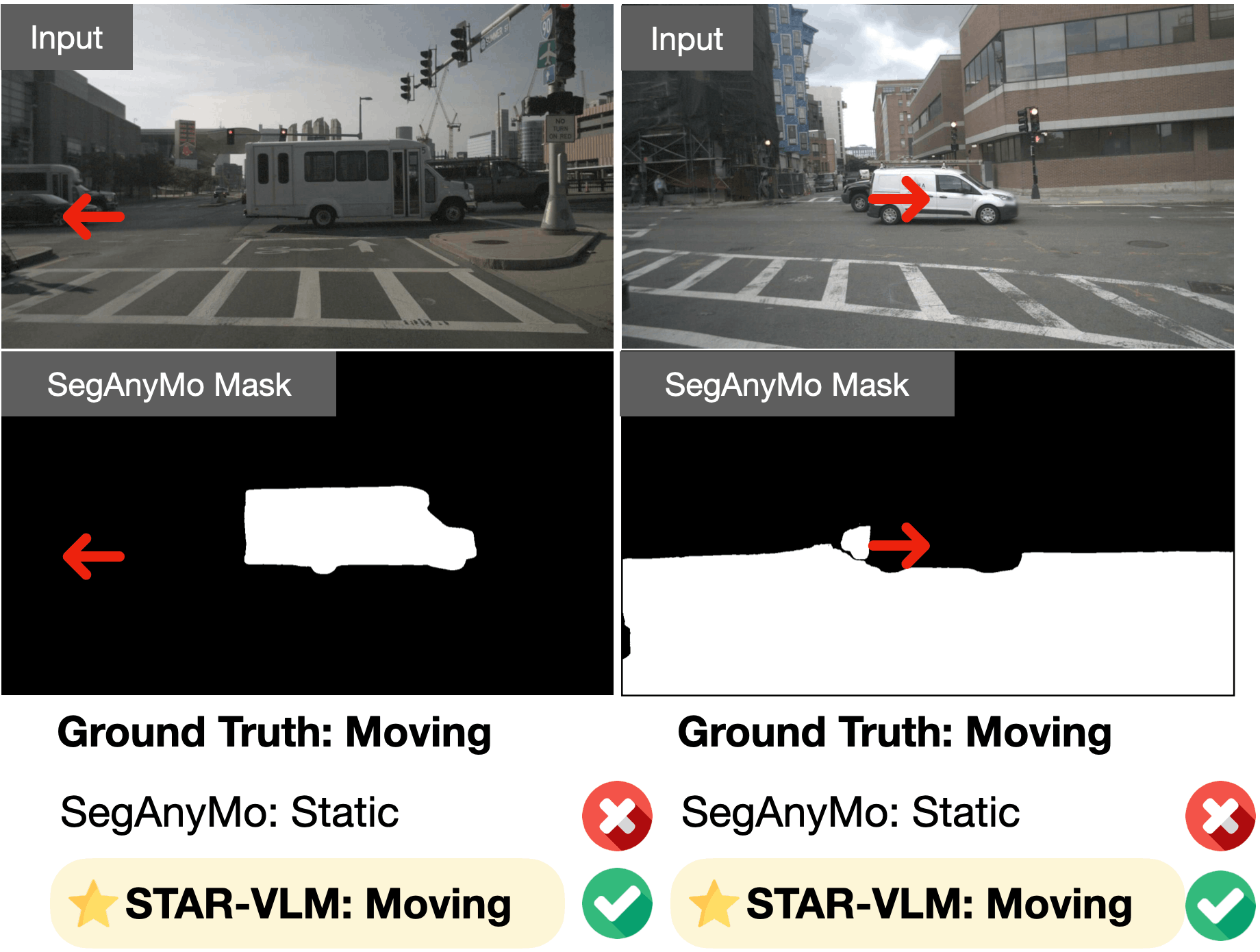}
    \vspace{-0.1in}
    \caption{Although SegAnyMo produces dense motion masks, it does not always accurately classify motion at the target pixels. By contrast, \algname reliably predicts the motion status of the object at the queried location.}
    \label{fig:seganymo}
\end{figure}

\begin{figure}
    \centering
    \includegraphics[width=1\linewidth]{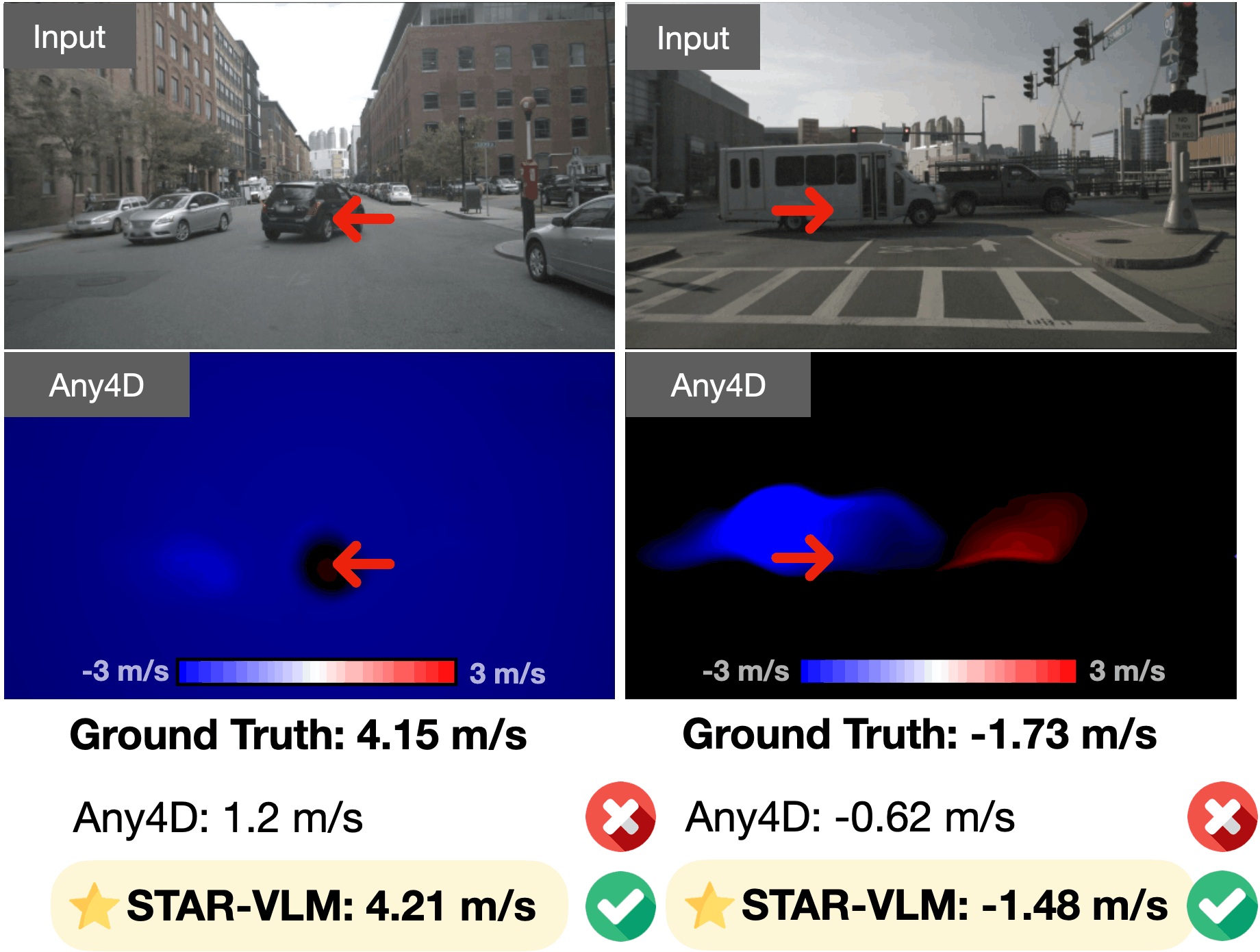}
    \vspace{-0.1in}
    \caption{Any4D produces dense scene flow, but the metric velocity estimation is off from ground truth. \algname accurately predicts the metric radial velocity of the object at the queried location.}
    \label{fig:any4d}
\end{figure}

\input{table/ablation}

\subsection{Results}
Table~\ref{tab:results} summarizes the experimental results. As qualitatively illustrated in Figure~\ref{fig:teaser}, Qwen2.5-VL and Qwen3-VL perform poorly on both motion classification and radial velocity estimation. Although Cosmos-Reason is developed for improved physical understanding, its performance on motion classification and metric velocity estimation remains limited. DepthLM, which is tailored for metric depth estimation, fails to produce meaningful motion classification results, likely because it is trained primarily to reason about and output depth values. Its radial velocity predictions are also inaccurate, likely due to its bias toward depth-centered prediction patterns.

For task-specific baselines, SegAnyMo~\cite{seganymotion} predicts dense pixel-level motion segmentation. We use the prediction at the query pixel as the output of the method. While SegAnyMo can generate dense and visually clear motion masks through SAM, we observe that it often captures only a single moving object in the scene and may fail when multiple moving objects overlap, as illustrated in Figure~\ref{fig:seganymo}. Overall, \algname achieves better motion classification accuracy than SegAnyMo when evaluated against both human annotations and radar measurements.

In contrast, Any4D~\cite{any4d} predicts dense pixel-level scene flow from consecutive frames. For evaluation, we extract the predicted per-pixel velocity and convert it into radial velocity. Although Any4D captures reasonable relative motion patterns across pixels, its metric velocity estimates remain less accurate. Consequently, \algname surpasses Any4D in metric velocity estimation, while requiring only radar measurements as training supervision, as shown in Figure~\ref{fig:any4d}.

\subsection{Cross-Task Ablation Studies}
\label{sec:cross-task}
Table~\ref{tab:ablation} shows the effect of incorporating ego-velocity estimation into joint training. In the single-task setting, the model trained jointly on radial velocity estimation and ego-velocity estimation outperforms the model trained only on radial velocity estimation, benefiting from the additional metric velocity supervision. In the joint-task setting, simply combining motion classification and radial velocity estimation leads to worse performance than single-task training, likely because the two tasks are not directly aligned and jointly learning them increases the optimization difficulty. However, adding ego-velocity estimation yields the best overall performance. A possible explanation is that ego-velocity supervision helps bridge motion classification and radial velocity estimation by introducing ego-motion awareness. This allows the model to implicitly learn ego-motion compensation, leading to improved joint-task training.
% , ego-velocity joint training improves motion classification accuracy. A possible explanation for this gain is discussed in Figure~\ref{fig:egovel_training}.

\section{Conclusion}

In this paper, we introduced \algname, a spatiotemporal VLM that leverages radar-derived supervision to improve motion and metric reasoning from video. Unlike existing spatiotemporal VLMs that often depend on human annotations or synthetic data, our approach uses Doppler information from low-cost automotive radar, a sensor already widely deployed in on-road vehicles, as a source of supervision, enabling annotation-free training.
Our experiments show that radar-derived motion labels and radial velocity serve as effective supervision signals for learning object motion and metric radial velocity estimation. Moreover, jointly training with ego-velocity estimation further improves both motion classification and radial velocity estimation, indicating beneficial cross-task transfer through shared spatiotemporal reasoning. These findings highlight radar as a scalable and annotation-efficient source of physically grounded supervision for VLM training in autonomous driving.
We further demonstrate that \algname achieves state-of-the-art performance on both motion classification and metric velocity estimation, outperforming even task-specific methods tailored to each task.
The current method is limited to relative radial velocity estimation. Future work could extend the method to full velocity or global velocity estimation and combine it with VLAs.

%% Use plainnat to work nicely with natbib. 
%\FloatBarrier
% \footnotesize
\small
{%\fontsize{7.9}{9.5}\selectfont
\bibliographystyle{IEEEtran}
\bibliography{main}

\end{document}

%% file: table/results.tex
\begin{table*}[t!]
\centering
\small
\setlength{\tabcolsep}{3pt}
\renewcommand{\arraystretch}{1.5}
\begin{tabular}{ccccccccclcc}
\hline
\multirow{3}{*}{\textbf{Model Types}} & \multirow{3}{*}{} & \multirow{3}{*}{\textbf{VLMS}} & \multirow{3}{*}{\textbf{Size}} & \multicolumn{5}{c}{\textbf{STAR-Bench}} & \multicolumn{1}{c}{} & \multicolumn{2}{c}{\textbf{Public Benchmark}} \\ \cline{5-9} \cline{11-12} 
 &  &  &  & \multicolumn{2}{c}{\textbf{Motion Classification Acc. ($\uparrow$)}} & \textbf{} & \multicolumn{2}{c}{\textbf{Radial Velocity MAE ($\downarrow$)}} & \multicolumn{1}{c}{} & \multicolumn{2}{c}{\textbf{TADBench}} \\ \cline{5-6} \cline{8-9} \cline{11-12} 
 &  &  &  & Anno. & Radar & \textbf{} & Anno. & Radar & \multicolumn{1}{c}{} & VQA ($\uparrow$) & MCQ ($\uparrow$) \\ \hline
\multirow{4}{*}{General VLM} &  & Qwen2.5-VL & 3B & 0.35 & 0.50 &  & 25.03 & 27.17 &  & 0.73 & 0.49 \\
 &  & Qwen2.5-VL & 7B & 0.44 & 0.65 &  & 20.30 & 26.84 &  & 0.10 & 0.57 \\
 &  & Qwen3-VL & 4B & 0.46 & 0.83 &  & N/A & N/A &  & 0.63 & 0.61 \\
 &  & Qwen3-VL & 8B & 0.54 & 0.62 &  & 16.47 & 26.30 &  & 0.64 & 0.66 \\ \hline
Physical VLM &  & Cosmos-Reason 1 & 7B & 0.44 & 0.63 &  & 10.42 & 11.77 & \multicolumn{1}{c}{} & 0.23 & 0.56 \\ \hline
\multirow{2}{*}{Metric-Depth VLM} &  & DepthLM & 3B & N/A & N/A &  & 23.16 & 26.33 &  & N/A & N/A \\
 &  & DepthLM-Pixtral & 12B & N/A & N/A &  & 21.12 & 27.92 &  & N/A & N/A \\ \hline
Ours &  & STAR-VLM & 4B & \textbf{0.80} & \textbf{0.94} &  & \textbf{1.94} & \textbf{1.37} &  & \textbf{0.81} & \textbf{0.72} \\ \hline
\multirow{2}{*}{Task-Specific Models} &  & SegAnyMo & -- & 0.73 & 0.77 &  & -- & -- & \multicolumn{1}{c}{} & -- & -- \\
 &  & Any4D & -- & -- & -- &  & 2.31 & 3.60 & \multicolumn{1}{c}{} & -- & -- \\ \hline
\end{tabular}
\caption{Comparison across base models, physical VLMs, metric-depth VLMs, and our variants on motion classification and radial velocity estimation. N/A means the model completely fails to output in the desired format. ``-" in the table means the model is not designed to do the task. The definition of Visual Question-Answer (VQA) and Multiple-Choice Question (MCQ) in TADBench are explained in Sec.~\ref{sec:tadbench}.}
\label{tab:results}
\end{table*}

%% file: table/ablation.tex
\begin{table*}[t!]
\centering
\small
\setlength{\tabcolsep}{6pt}
\renewcommand{\arraystretch}{1.3}
\begin{tabular}{ccccccccc}
\hline
\multicolumn{3}{c}{\textbf{Training Tasks}} & \textbf{} & \multicolumn{2}{c}{\multirow{2}{*}{\textbf{Motion Classification Accuracy ($\uparrow$)}}} &  & \multicolumn{2}{c}{\multirow{2}{*}{\textbf{Radial Velocity MAE ($\downarrow$)}}} \\ \cline{1-3}
\multirow{2}{*}{\textbf{Motion Classification}} & \multirow{2}{*}{\textbf{Radial Velocity}} & \multirow{2}{*}{\textbf{Ego Velocity}} &  & \multicolumn{2}{c}{} & \textbf{} & \multicolumn{2}{c}{} \\ \cline{5-6} \cline{8-9} 
 &  &  &  & Anno. & Radar & \textbf{} & Anno. & Radar \\ \hline
\multicolumn{9}{c}{Single-Task (Motion Classification or Radial Velocity) Training} \\ \hline
\checkmark &  &  &  & 0.79 & 0.93 &  & -- & -- \\
 & \checkmark &  &  & -- & -- &  & 2.23 & 1.42 \\
 & \checkmark & \checkmark &  & -- & -- &  & \textbf{1.91} & \textbf{1.20} \\ \hline
\multicolumn{9}{c}{Joint Motion Classification and Radial Velocity Training} \\ \hline
\checkmark & \checkmark &  &  & 0.62 & 0.81 &  & 4.91 & 3.51 \\
\checkmark & \checkmark & \checkmark &  & \textbf{0.8} & \textbf{0.94} &  & \textbf{1.89} & \textbf{1.20} \\ \hline
\end{tabular}
\caption{Cross-task ablation studies show that joint training with ego-velocity estimation improves both motion classification and radial velocity prediction.}
\vspace{-0.15in}
\label{tab:ablation}
\end{table*}